\documentclass[11pt]{article}

\usepackage[preprint]{acl}

\usepackage{times}
\usepackage{latexsym}

\usepackage[T1]{fontenc}

\usepackage[utf8]{inputenc}

\usepackage{microtype}

\usepackage{inconsolata}

\usepackage{graphicx}
\usepackage{booktabs}
\usepackage{multirow}
\usepackage{algorithm}
\usepackage{algorithmic}
\usepackage{amssymb}
\usepackage{amsmath}
\usepackage[table]{xcolor}
\title{From Implicit to Explicit: Enhancing Self-Recognition in Large Language Models}

\author{
 \textbf{Yinghan Zhou\textsuperscript{1}},
 \textbf{Zhuwei Feng\textsuperscript{1}},
 \textbf{Juan Wen\textsuperscript{1}}\thanks{Corresponding author},
 \textbf{Wanli Peng\textsuperscript{1}}\thanks{Corresponding author},
 \textbf{Zhengxian Wu\textsuperscript{1}},
 \textbf{Yiming Xue\textsuperscript{1}}\\
 \textsuperscript{1}College of Information and Electrical Engineering, China Agricultural University.
\\
 \small{
\href{zhouyh@cau.edu.cn}{zhouyh@cau.edu.cn},
\href{y146776149@gmail.com }{y146776149@gmail.com },
\href{wenjuan@cau.edu.cn}{wenjuan@cau.edu.cn},
\href{wlpeng@cau.edu.cn}{wlpeng@cau.edu.cn},}\\
 \small{
\href{wzxian@cau.edu.cn}{wzxian@cau.edu.cn},
\href{xueym@cau.edu.cn}{xueym@cau.edu.cn}
 }}

\begin{document}
\maketitle
\begin{abstract}
Large language models (LLMs) have been shown to possess a degree of self-recognition ability, which used to identify whether a given text was generated by themselves. 
Prior work has demonstrated that this capability is reliably expressed under the pair presentation paradigm (PPP), where the model is presented with two texts and asked to choose which one it authored. 
However, performance deteriorates sharply under the individual presentation paradigm (IPP), where the model is given a single text to judge authorship.  
Although this phenomenon has been observed, its underlying causes have not been systematically analyzed. 
In this paper, we first investigate the cause of this failure and attribute it to implicit self-recognition (ISR). 
ISR describes the gap between internal representations and output behavior in LLMs: under the IPP scenario, the model encodes self-recognition information in its feature space, yet its ability to recognize self-generated texts remains poor.
To mitigate the ISR of LLMs, we propose cognitive surgery (CoSur), a novel framework comprising four main modules: representation extraction, subspace construction, authorship discrimination, and cognitive editing. 
Experimental results demonstrate that our proposed method improves the self-recognition performance of three different LLMs in the IPP scenario, achieving average accuracies of 99.00\%, 97.69\%, and 97.13\%, respectively.
\end{abstract}

\section{Introduction}
It has recently been found that large language models
(LLMs) possess self-recognition ability, enabling them
to distinguish their own writing ("self-texts") from that of
humans and other models ("other-texts") \citep{panickssery2024llm,ackerman2025inspection}.
This finding raises concerns, as a prior study \citep{panickssery2024llm} found that LLMs with self-recognition ability often exhibit a self-preference bias when acting as judges.
Such a bias can compromise the reliability of LLM-based evaluation and decision-making.
On the positive side, the self-recognition ability of LLMs can be leveraged in future defenses against malicious prompting, for example, by enabling models to detect externally introduced instructions that conflict with the model’s own prior outputs.

To assess the self-recognition ability of LLMs, researchers designed two different paradigms: the pair presentation paradigm (PPP) and the individual presentation paradigm (IPP). 
In the PPP scenario, the model is shown two texts: one generated by the model being tested and the other by either a human or another model. 
The model is then asked to identify which of the two texts was generated by itself. 
In the IPP scenario, the model is shown a single text and asked to determine whether it was generated by itself. 
A previous study \citep{panickssery2024llm} revealed that in the PPP scenario, the model demonstrated strong self-recognition ability. 
However, in the IPP scenario, the LLM's self-recognition ability diminished. 
As reported in Table 1 of \citep{ackerman2025inspection}, the base LLM achieved prediction accuracies below 50.3\% across four datasets.

To investigate this phenomenon, we extract the last-token hidden representations at the final layer of the model for self-texts and other-texts under the IPP scenario.
As shown in Figure \ref{confusion_matrix}, a logistic regression (LR) classifier trained on these representations achieves over 90\% classification accuracy, indicating strong linear separability between self-text and other-text representations.
This reveals a gap between internal representations and output behavior: while the model encodes self-recognition information in the feature space under the IPP scenario, its self-recognition performance remains poor, which is referred to as implicit self-recognition (ISR).

To enhance the self-recognition capability of LLMs under the IPP scenario, we conduct an in-depth analysis of ISR.
As demonstrated in Table \ref{feature}, we find that although the distributions of these representations are highly similar, the pairwise similarity relationships among samples differ across sources. 
Building on this insight, we raise a new question: \textit{Is it possible to enhance LLM performance in the IPP scenario by mitigating ISR?}

In this paper, we propose a novel method named cognitive surgery (CoSur) to mitigate the ISR, enhancing the self-recognition capability of LLMs in IPP scenarios.
The CoSur consists of four modules: representation extraction, subspace construction, authorship discrimination, and cognitive editing. 
Specifically, we first extract hidden representations of self-texts and other-texts from the LLM under the IPP scenario. 
We then apply singular value decomposition (SVD) separately to these representations to capture their structural characteristics. 
The leading right singular vectors are used to construct the self-recognition subspace and the other-recognition subspace, respectively.
Given a text, we extract the hidden representation of its last token from the final layer of the LLM in the IPP scenario.
And then, we perform authorship discrimination by computing its representation projection energy onto the self-recognition and other-recognition subspaces.
Finally, we design the cognitive editing to induce the LLM to generate the correct response. 
Extensive experiments on three LLMs demonstrate that CoSur effectively enhances their self-recognition capability in the IPP scenario.
Our contributions are summarized as follows:
\begin{itemize}
    \item We identify the implicit self-recognition (ISR) in LLMs, revealing the gap between distinct feature-level separability and poor output-level performance for self-texts and other-texts under the IPP scenario.
    \item We comprehensively analyze the reason why the LLMs have the ISR Phenomenon in the IPP scenario. Moreover, we propose cognitive surgery (CoSur) to enhance self-recognition of LLMs under the IPP scenario.
    \item  Experiments demonstrate that our proposed method enhances the performance across three different LLMs in the IPP scenario, achieving average accuracy of 99.00\%, 97.69\%, and 97.13\%, respectively.
\end{itemize}

\section{Related Work}
\subsection{self-recognition ability of LLMs}
The self-recognition ability of LLM refers to their capacity to identify texts they have generated \cite{NEURIPS2024_75377263,laine2023towards,wang2024mm,ajeyasitu}. 
\citet{panickssery2024llm} reported that several LLMs, including Llama2-7b-chat, demonstrated out-of-the-box (without fine-tuning) self-recognition capabilities using a summary writing and recognition task.
\citet{laine2024me} used more challenging text continuation and recognition tasks to demonstrate self-recognition abilities in LLMs. It highlighted how task success could be elicited with different prompts and across different models.
\citet{ackerman2025inspection} found that the Llama3-8b-Instruct succeeded at self-recognition across diverse tasks, whereas the base model performed poorly, especially in the IPP scenario. They also extracted a "self-recognition" vector in the residual stream, allowing users to steer the LLM to claim or disclaim authorship during generation and to believe or disbelieve that it had written arbitrary texts when reading them.
These studies demonstrated that LLMs possess self-recognition ability.

\subsection{Representation Editing}
Representation editing is a class of techniques that directly manipulate the latent representations of a model to improve its performance and align it with desired attributes \cite{kong2024aligning,wu2024advancing}.
\citet{liang2024controllable} found that representation editing could control aspects of text generation, such as safety, sentiment, thematic consistency, and linguistic style.
\citet{adila2024free} used embedding editing for general, rather than personalized, alignment to broad human preferences, relying on self-generated synthetic data.
\citet{wu2024reft} showed that the representation editing can even surpass fine-tuning-based methods by intervening on hidden representations within the linear subspace defined by a low-rank projection matrix. 
Inspired by representation editing, as long as the true authorship of the text is determined, we can directly manipulate the LLM's hidden representations to generate the correct response.
Based on this, we propose CoSur, the details of which will be introduced in the section \ref{'CoSur'}.

\begin{figure}[t]
\centering
\includegraphics[width=0.48\textwidth]{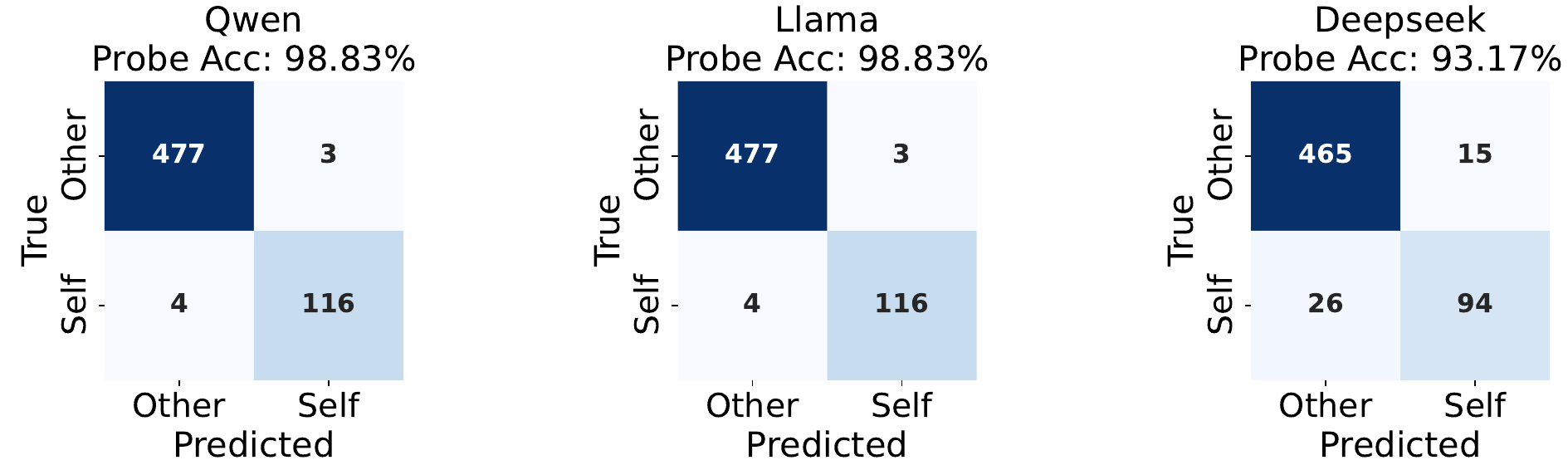} 
\caption{Evaluation of LLM self-recognition capabilities via linear probing across three models.}
\label{confusion_matrix}
\end{figure}

\begin{figure*}
\centering
\includegraphics[width=\textwidth]{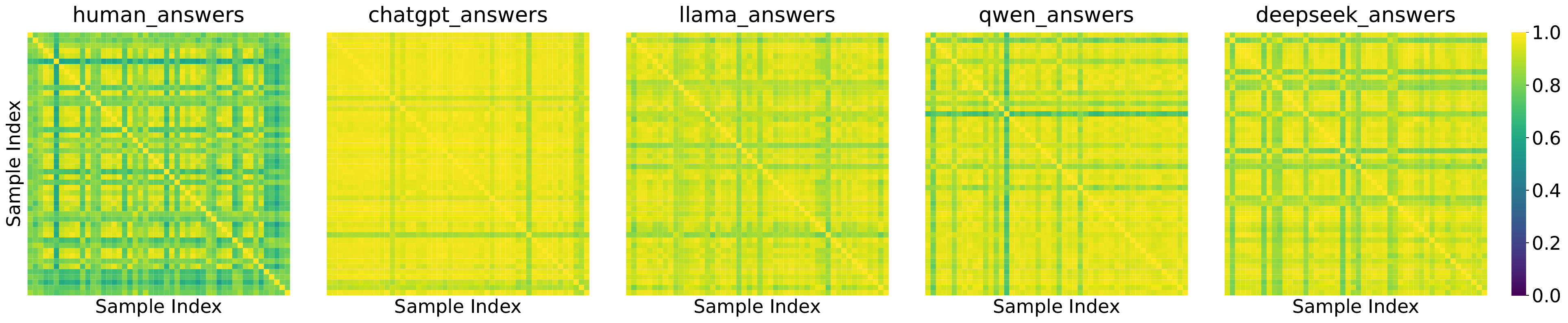} 
\caption{Pairwise Representation Similarity Heatmap on Qwen.}
\label{RSA_heatmap}
\end{figure*}

\section{The Implicit Self-Recognition of LLMs}

Previous studies have demonstrated that LLMs possess self-recognition ability.
However, in the IPP scenario, when the LLM is asked about the authorship of a single text, this ability is not reflected in its response.
To investigate the internal mechanisms underlying self-recognition in LLMs, we randomly sample 1000 questions from the HC3 dataset \citep{guo2023closeChatGPThumanexperts}, covering five domains: open-domain knowledge, finance, medicine, law, and psychology.
These questions are used to generate responses from three LLMs: Qwen3-8B (Qwen)\citep{qwen3technicalreport}, Llama3.1-8B (Llama)\citep{Llama3modelcard}, and Deepseek-R1-0528-Qwen3-8B (Deepseek) \citep{Deepseek-r1-0528}. 
For each model, we extract the final-layer hidden representation of the last token for texts generated by different sources in the IPP scenario. 
The detailed prompt used in the IPP scenario is provided in Appendix \ref{app:prompt}.
We train a linear probe using logistic regression on these representations to distinguish self-generated from other-generated text. The probe attains classification accuracies exceeding 90\% across all three models, as shown in Figure \ref{confusion_matrix}. These results demonstrate strong linear separability between the representations of self- and other-generated texts.
This reveals a gap between internal representation and output behavior: while the model encodes discriminative information in feature space, it fails to express it in its outputs. 
We regard this phenomenon as the implicit self-recognition (ISR) of LLMs.

Empirically, we quantify the distributional difference between representations of texts from different sources using three metrics, including cosine similarity (CS), Maximum Mean Discrepancy (MMD), and Centered Kernel Alignment (CKA).
The definitions of these metrics are provided in Appendix \ref{app:distribution}.
As shown in Table \ref{feature}, these features exhibit consistently high CS and low MMD, indicating that they are highly similar in feature space. 
\begin{table}[t]
\centering
\setlength{\tabcolsep}{2mm}
\begin{tabular}{cccc}
\toprule
      & CS & MMD & CKA  \\
\midrule
    Qwen-human & 0.9293 & 0.0047 & 0.0880 \\
    Qwen-ChatGPT & 0.9731 & 0.0160 & 0.0623 \\ 
    Qwen-Llama & 0.9494 & 0.0066 & 0.0332 \\
    Qwen-Deepseek & 0.9841 & 0.0055 & 0.1035 \\
\bottomrule
\end{tabular}
\caption{The distance between feature representations of texts from different sources in Qwen.
‘A–B’ denotes the distance between category A and category B. 
}
\label{feature}
\end{table}
In contrast, the CKA scores between these representation sets, which evaluate the structural similarity in the feature space by calculating the similarity between the Gram matrices of two feature matrices,  are notably low.
The experimental results reveal that although the features are close in space, the internal structure of the representations differs significantly across sources. 
To visualize these differences, we randomly selected 50 responses generated by different models and computed a pairwise feature similarity heatmap.
As shown in Figure \ref{RSA_heatmap}, the internal structure of representations exhibits significant variation across texts from different sources.
These results demonstrate that the LLM effectively extracts the features relevant to self-recognition under the IPP scenario.

We attribute the phenomenon to the information loss during the mapping from the feature space to the discrete vocabulary space in the LLM.
Specifically, the output probability distribution $\mathbf{P}$ is obtained via a linear projection followed by a softmax function:

\begin{equation}
    \mathbf{P} = softmax (\mathbf{W} \mathbf{h} + \mathbf{b}),
\end{equation}
where $softmax( \cdot)$ denotes the softmax function, $\mathbf{W} \in \mathbb{R}^{|V| \times d}$ is the output projection matrix, $\mathbf{b} \in \mathbb{R}^{|V|}$ is the bias term, and $|V|$ is the vocabulary size. 
This transformation constrains the representation to a task-specific decision manifold defined by the vocabulary simplex and the model’s training objective.
Therefore, only those components of $\mathbf{h}$  that align with token-level decision boundaries are effectively expressed at the output level, while other internally encoded signals, such as those distinguishing self-generated texts, are filtered out.
From an information-theoretic perspective, we quantify the discrepancy between internal representations and output expressions using Mutual Information (MI).
The mapping from $\mathbf{h}$ to $\mathbf{P}$ forms a Markov chain $y \rightarrow \mathbf{h}\rightarrow \mathbf{P}$. where $y$ denotes the source label of a given text.
By the Data Processing Inequality (DPI), the mutual information between the output distribution and $y$ is upper-bounded by that of the hidden representation. 
That is, $I(y; \mathbf{P}) \le I(y; \mathbf{h}).$
This upper bound formalizes the ISR phenomenon: although self-recognition signals are encoded in the hidden representations $\mathbf{h}$, they fail to pass through the bottleneck to the output probabilities $\mathbf{P}$.
Our findings align with the information bottleneck theory \cite{tishby2000information}, which is also demonstrated on other LLMs, including Llama and Deepseek. 
The details can be found in Appendix B.

\begin{figure*}[t]
\centering
\includegraphics[width=0.9\textwidth]{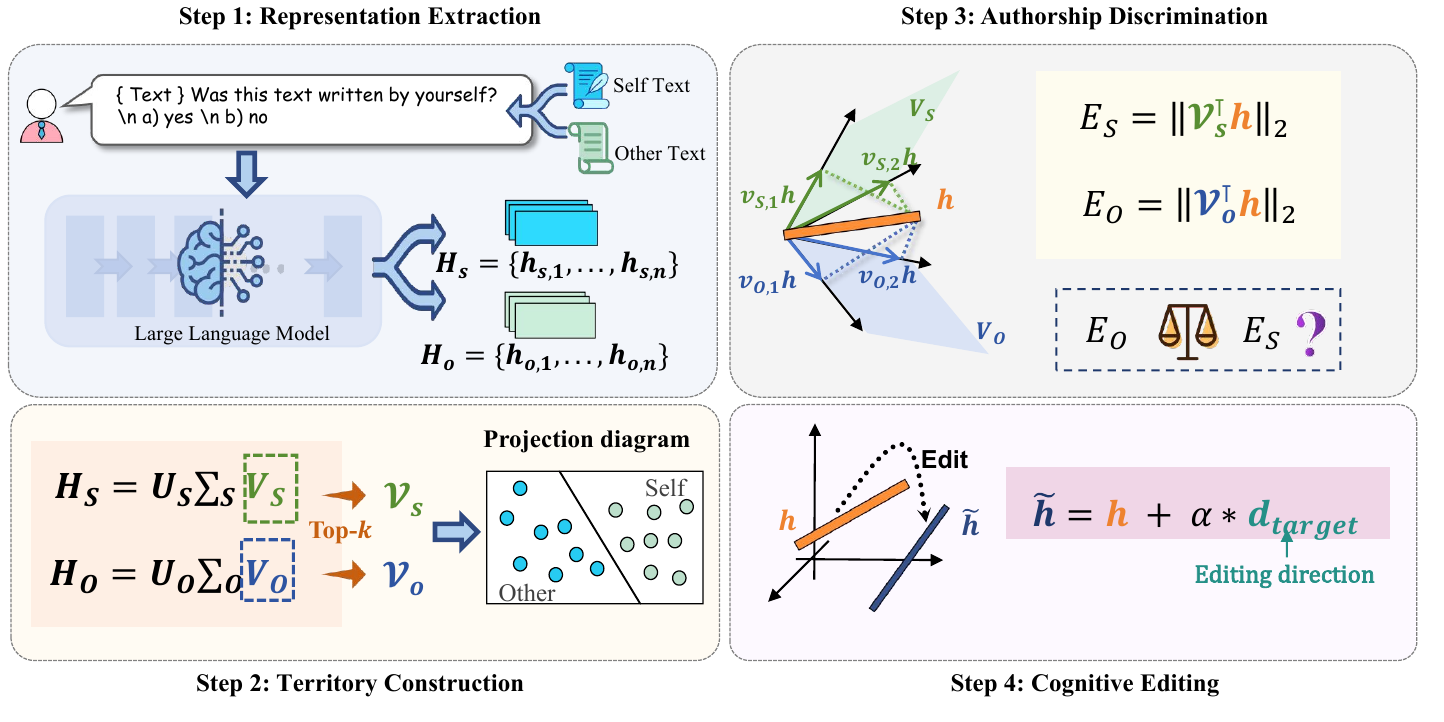} 
\caption{The framework of CoSur. 
\textbf{Step 1 (Representation Extraction):} 
Extract the final-layer representations of self-generated and other-generated texts from the LLM, denoted as $\mathbf{H}_s$ and $\mathbf{H}_o$, respectively.
\textbf{Step 2 (Subspace Construction):} SVD is applied to $\mathbf{H}_s$ and $\mathbf{H}_o$ to construct the subspace for each text category, denoted as $\mathbf{V}_s$ and $\mathbf{V}_s$, respectively.
\textbf{Step 3 (Authorship Discrimination):} For a given sample $t$, compute its projection energy onto $\mathbf{V}_s$ and $\mathbf{V}_s$ to infer the authorship of $t$.
\textbf{Step 4 (Cognitive Editing):} Edit the feature representation $\mathbf{h}$ to approach the target response, denoted as $\tilde{h}$, thereby promoting the LLM to generate the correct reply.
}
\label{framework}
\end{figure*}

\section{CoSur}
\label{'CoSur'}
To further bridge the gap between hidden representations and output behavior, we propose Cognitive Surgery (CoSur), which framework that is shown in Figure \ref{framework}.
It consists of four modules: representation extraction, subspace construction, authorship discrimination, and cognitive editing.

\subsection{Representation Extraction}
Let $\mathbf{T_s} = \{ \mathbf{t}_{s,1}, \mathbf{t}_{s,2}, \ldots, \mathbf{t}_{s,N} \}$ denotes a set of texts generated by the LLM itself, and $\mathbf{T_o} = \{ \mathbf{t}_{o,1}, \mathbf{t}_{o,2}, \ldots, \mathbf{t}_{o,N} \}$ represents a set of texts from other source. 
Under the IPP scenario, each text $\mathbf{t}_{s, i} \in \mathbf{T_s}$ and $\mathbf{t}_{o, i} \in \mathbf{T_o}$ is independently fed into the LLM to extract the hidden representation of its last token from the final layer of the LLM, denoted as $\mathbf{h_{s, i}} \in \mathbb{R}^d$ and $\mathbf{h_{o, i}} \in \mathbb{R}^d$, respectively:

\begin{equation}
\mathbf{h}_{s,i} = LLM (\mathbf{t}_{s,i}), \quad \mathbf{h}_{o,i} = LLM (\mathbf{t}_{o,i})
\end{equation}
The representations for each category are stacked to form the sets 
$\textbf{H}_{\text{s}} \in \mathbb{R}^{N \times d}$
and
$\textbf{H}_{\text{o}} \in \mathbb{R}^{N \times d}$, respectively.

\subsection{Subspace Construction}


We apply SVD to extract the most discriminative components between categories.

\begin{equation}
\mathbf{H_s} = \mathbf{U_s} \mathbf{\Sigma_s} \mathbf{V_s}^\top, \quad \mathbf{H_o} = \mathbf{U_o} \mathbf{\Sigma_o} \mathbf{V_o}^\top
\end{equation}
where $\mathbf{U_s}\in \mathbb{R}^{N \times N}$ and $\mathbf{U_o} \in \mathbb{R}^{N \times N}$ are the left singular matrices.
$\mathbf{\Sigma_s} \in \mathbb{R}^{N \times d}$ and $\mathbf{\Sigma_o} \in \mathbb{R}^{N \times d}$ are diagonal matrices containing the singular values.
$\mathbf{V_s^\top} \in \mathbb{R}^{d \times d}$ and $\mathbf{V_o^\top} \in \mathbb{R}^{d \times d}$ are the right singular matrices.
We then extract the top-$k$ right singular vectors from \( \mathbf{V_s} \) and \( \mathbf{V_o} \) to serve as the basis vectors defining the self-recognition subspace $\mathbf{\mathcal{V}_s} \in \mathbb{R}^{d\times k}$ and the other-recognition subspace $\mathbf{\mathcal{V}_o} \in \mathbb{R}^{d\times k}$.
We conduct experiments to examine the impact of the choice of $k$ on the results, as detailed in section \ref{ab}.

To evaluate the separability of these subspaces, we compute the Normalized Grassmann Distance (NGD) and Normalized Frobenius Distance (NFD) between them. 
As shown in Table \ref{subspace distance}, the NGD and NFD between these subspaces are large, indicating significant divergence between their corresponding subspaces. The definitions of the metrics and their measurement results on other LLMs are provided in Appendix \ref{app: subspace}.

\begin{table}[t]
\centering
\setlength{\tabcolsep}{2mm}
\begin{tabular}{ccc}
\toprule
      Other Source & NGD & NFD   \\
\midrule
    Human & 0.7449 & 0.8633\\
    ChatGPT & 0.6590 & 0.8633 \\
    Deepseek & 0.6458 & 0.7424 \\
    Llama  & 0.6458 & 0.7774 \\ 
\bottomrule
\end{tabular}
\caption{Measurement of distances between different subspaces using Qwen.
The self-recognition subspace is constructed from Qwen-texts, and the Other-source column indicates the model used to generate texts for constructing the other-recognition subspace.
}
\label{subspace distance}
\end{table}

\begin{table*}[t]
\centering
\setlength{\tabcolsep}{1mm}
\begin{tabular}{cc|ccccccccccccccc}
\toprule
    \multirow{2}{*}{Model}    & \multirow{2}{*}{Other Source} & \multicolumn{2}{c}{Base} & \multicolumn{2}{c}{ICL} & \multicolumn{2}{c}{LoRa-FT} & \multicolumn{2}{c}{CoSur$_{LR}$}            & \multicolumn{2}{c}{CoSur}         \\
                          &                               & ACC         & F1         & ACC        & F1         & ACC          & F1           & ACC                  & F1                   & ACC             & F1              \\ \midrule
\multirow{4}{*}{Qwen}     & Human                         & 26.25       & 25.91      & 56.50      & 54.59      & 97.25        & 97.25        & 97.25                & 97.25                & \textbf{100.00} & \textbf{100.00} \\
                          & chatgpt                       & 25.00       & 24.98      & 48.00      & 46.54      & 99.00        & 98.99        &  \underline{ 99.25}          &  \underline{ 99.25}          & \textbf{99.75}  & \textbf{99.75}  \\
                          & Llama                         & 29.47       & 29.19      & 39.80      & 37.75      & 95.47        & 95.45        & \textbf{100.00}      & \textbf{100.00}      &  \underline{ 99.25}     &  \underline{ 99.25}     \\
                          & Deepseek                      & 12.00       & 11.43      & 41.00      & 37.39      & 90.75        & 90.68        &  \underline{ \textbf{98.75}} &  \underline{ \textbf{98.75}} &  \underline{ 97.00}     &  \underline{ 97.00}     \\ \midrule
\multirow{4}{*}{Llama}    & Human                         & 13.60       & 13.58      & 37.28      & 27.47      & 49.62        & 33.17        &  \underline{ 94.25}          &  \underline{ 94.25}          & \textbf{96.50}  & \textbf{96.50}  \\
                          & chatgpt                       & 10.83       & 10.48      & 30.23      & 23.96      & 62.72        & 56.98        & \textbf{98.25}       & \textbf{98.26}       &  \underline{ 98.00}     &  \underline{ 98.00}     \\
                          & Deepseek                      & 12.85       & 12.31      & 31.99      & 26.45      & 74.81        & 73.28        & \textbf{99.25}       & \textbf{99.25}       &  \underline{ 97.25}     &  \underline{ 97.24}     \\
                          & Qwen                          & 12.85       & 12.49      & 31.74      & 26.50      & 74.31        & 72.65        & \textbf{100.00}      & \textbf{100.00}      &  \underline{ 99.00}     &  \underline{ 99.00}     \\ \midrule
\multirow{4}{*}{Deepseek} & Human                         & 9.25        & 9.24       & 20.00      & 18.28      &  \underline{ 89.25}  &  \underline{ 89.25}  & 88.00                & 87.84                & \textbf{99.50}  & \textbf{99.50}  \\
                          & chatgpt                       & 9.25        & 9.24       & 37.25      & 30.22      & 93.00        & 92.98        &  \underline{ 97.50}          &  \underline{ 97.50}          & \textbf{98.00}  & \textbf{98.00}  \\
                          & Llama                         & 12.59       & 12.41      & 7.05       & 6.90       & 55.67        & 45.28        & \textbf{98.50}       & \textbf{98.49}       &  \underline{ 97.50}     &  \underline{ 97.49}     \\
                          & Qwen                          & 12.50       & 12.32      & 38.75      & 30.34      & 53.00        & 39.67        & \textbf{97.25}       & \textbf{97.25}       &  \underline{ 93.50}     &  \underline{ 93.29}     \\ \midrule \rowcolor{blue!10}
\multicolumn{2}{c|}{Average}                               & 15.54       & 15.30      & 34.97      & 30.53      & 77.90        & 73.80        &  \underline{ 97.35}          &  \underline{ 97.34}          & \textbf{97.94}  & \textbf{97.92}  \\ \bottomrule
\end{tabular}
\caption{Performance of three LLMs in the IPP scenarios.
“Other Source” denotes the generation source of the other-texts, while self-texts are generated by the evaluated LLM itself.
Bold and underlined values denote the best and second-best results, respectively.}
\label{main}
\end{table*}

\begin{table}[t]
\centering
\setlength{\tabcolsep}{0.5mm}
\begin{tabular}{cc|ccc}
\toprule
Model   & Other Source & ICL & LoRa-FT     & CoSur      \\ \midrule
\multirow{2}{*}{Qwen}     & Llama                         & 51.39      &  \textbf{ 82.37}    & \underline{70.75}              \\
                          & Deepseek                      & 49.00      & \underline{51.75}          &  \textbf{ 59.50}    \\ \midrule
\multirow{2}{*}{Llama}    & Deepseek                      & 30.23      & 49.87          & \textbf{70.75}  \\
                          & Qwen                          & 31.99      & 50.38          & \textbf{81.25} \\  \midrule
\multirow{2}{*}{Deepseek} & Llama                         & 25.75      & \textbf{90.43} & 71.25 \\
                          & Qwen                          & 25.75      & \textbf{50.75} &  \underline{ 49.75}   \\  \midrule \rowcolor{blue!10}
\multicolumn{2}{c|}{Average}                               & 35.69      &  \underline{62.59}    & \textbf{67.21}\\ \bottomrule
\end{tabular}
\caption{Generalization accuracy for self-recognition across three LLMs in IPP scenarios, with source texts consisting of self-texts and ChatGPT-texts.}
\label{generalization}
\end{table}

\subsection{Authorship Discrimination}
In the reference stage, we introduce projection energy $E$ to quantify the intensity of text representations projected onto each subspace, which serves to determine the authorship of a given text. For a given text sample, the last token feature vector $\mathbf{h}$ from the final layer is extracted, and its projection energy $E_s$ and $E_o$ onto $\mathbf{V_s}$ and $\mathbf{V_o}$ are computed to infer the authorship of $t$.
\begin{equation}
E_s = \|\mathbf{ V_s}^\top \mathbf{h} \|_2, \quad E_o= \| \mathbf{V_o}^\top \mathbf{h} \|_2
\end{equation}
where $\| \cdot \|_2$ denotes the Euclidean norm.

Finally, the authorship of the text $t$ is determined by comparing its projection energies $E_s$ and $E_o$  onto the respective subspaces.

\begin{equation}
O(t)= \left\{
\begin{array}{ll}
s, & if \quad E_s > E_o\quad ,\\
o, & otherwise
\end{array} \right.
\label{onership}
\end{equation}
where $O(t)$ represents the authorship of $t$.

\subsection{Cognitive Editing}
To guide the LLM toward producing the desired response, we first identify the target tokens $tok_s$ and $tok_o$.
The vectors in the LLM's output projection corresponding to the two target tokens, denoted $\mathbf{w}_s \in \mathbb{R}^d$ and $\mathbf{w}_o \in \mathbb{R}^d$, are obtained from the LLM.
We then normalize the two weight vectors as follows:
\begin{equation}
\tilde{\mathbf{w}}_o= \frac{\mathbf{w}_o}{\|\mathbf{w}_o\|}, \quad
\tilde{\mathbf{w}}_s = \frac{\mathbf{w}_s}{\|\mathbf{w}_s\|}
\end{equation}

The target direction $\mathbf{d_{target}}$ is determined according to the value of $O(t)$:
\begin{equation}
\mathbf{d_{target}}= \left\{
\begin{array}{ll}
\tilde{\mathbf{w}}_o, & if \quad O(t)=o \quad ,\\
\tilde{\mathbf{w}}_s, & if \quad O(t)=s
\end{array} \right.
\label{direction}
\end{equation}

The hidden representation $\mathbf{h}$ is steered toward the target direction to obtain the edited representation $\mathbf{\tilde{h}}$, thereby facilitating the LLM to output the target token.

\begin{equation}
\mathbf{\tilde{h}}=\mathbf{h}+\alpha \cdot \mathbf{d_{target}}
\end{equation}
where $\alpha$ represents the editing strength hyperparameter.
We also conduct experiments to examine the impact of the choice of $\alpha$ on the results, as detailed in section \ref{ab}.
The complete algorithmic procedure is detailed in Appendix \ref{app:algorithm}.


\section{Experiments}

\subsection{Dataset}
We construct our dataset based on HC3 \citep{guo2023closeChatGPThumanexperts}, a large-scale question–answer corpus that includes responses from both human experts and ChatGPT across multiple domains, such as open-domain knowledge, finance, medicine, law, and psychology.
Specifically, we randomly sample 1,000 questions and use them to generate responses from three LLMs: Qwen3-8B (Qwen) \citep{qwen3technicalreport}, Llama3.1-8B (Llama) \citep{Llama3modelcard}, and Deepseek-R1-0528-Qwen3-8B (Deepseek) \citep{Deepseek-r1-0528}.
The resulting dataset is split into training, validation, and test sets with a ratio of 6:2:2.

\subsection{Baselines}
We conduct comparisons with four baseline methods to evaluate the effectiveness of our approach.
\textbf{(1) Base.}
The model directly predicts the authorship of the input text.
\textbf{(2) In-Context Learning (ICL).}
The LLM is provided with three self-generated texts and three texts generated by the other source as in-context examples.
\textbf{(3) Low-Rank Adaptation Fine-tuning (LoRA-FT).}
The LLM is fine-tuned using LoRA on a dataset consisting of 600 self-generated texts and 600 texts produced by the other source.
\textbf{(4) CoSur\textsubscript{LR}.}
A logistic regression (LR) classifier is used for authorship discrimination, and cognitive editing is applied to steer the hidden representations.

\subsection{Experimental Setting}
We evaluate the performance of CoSur on three different LLMs, including Qwen, Llama, and Deepseek. 
We select the top-$k$ right-singular vectors to construct the subspace ($k=8$)  and set the editing strength $\alpha=100$.
All experiments are run on two NVIDIA 4090 GPUs.
Additionally, we use stricter evaluation metrics, where accuracy (ACC) and F1 score (F1) are computed based solely on the output of LLMs, rather than the probability of the target token.

\subsection{Experimental Results}
\textbf{Self-recognition performance.}
We evaluate the effectiveness of CoSur on three different mainstream LLMs. 
As shown in Table \ref{main}, CoSur improves performance in most settings, achieving the highest average results in the IPP scenarios. Specifically, it outperforms the Base by 82.41\%. 
These results suggest that CoSur effectively mitigates the ISR exhibited by LLMs.
Although LoRA-FT achieves an average performance of 77.90\%, its results are inconsistent across different sources. 
This instability is due to semantic distribution shifts: even within a single source, the diversity of content makes it difficult for LoRA-FT to learn a reliable mapping to fixed labels.
CoSur\textsubscript{LR} consistently achieves higher performance across all other-text sources, reflecting that linear separability of self-recognition features in hidden representations is sufficient for robust classification.
To further investigate the causes of self-recognition failures in LLMs, we also evaluate the False Negative Rate (FNR).
A detailed analysis is provided in the Appendix \ref{app: additional_exp}.
Additionally, we present a case study comparing the top-10 tokens and their logit changes before and after editing, with detailed results provided in Appendix \ref{app:case}.

\begin{figure}[t]
\centering
\includegraphics[width=0.45\textwidth]{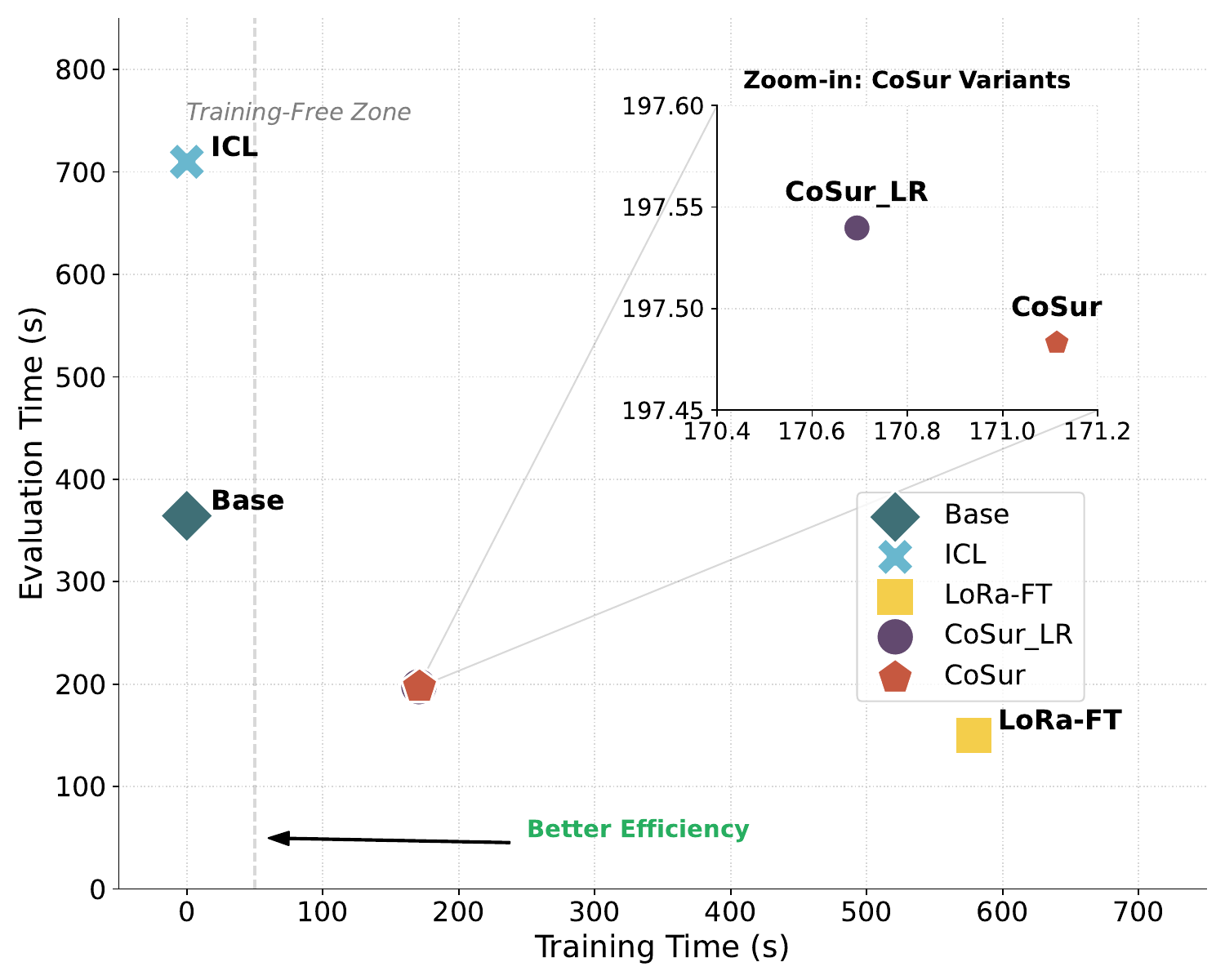} 
\caption{Average accuracy variation with different editing Strength $\alpha$ using Qwen.}
\label{efficiency}
\end{figure}

\textbf{Evaluation of generalization.} 
We evaluate the generalization capability of three LLMs in IPP scenarios across unseen text sources. 
For each target model, the other-recognition subspace is constructed using ChatGPT-generated texts.
Generalization is then evaluated by measuring recognition accuracy when the other-texts are generated by previously unseen models.
As shown in Table \ref{generalization}, CoSur achieves the highest average accuracy of 67.21\%, outperforming LoRA-FT and ICL. 
These results demonstrate that CoSur effectively leverages the intrinsic structure of the self-recognition subspace to maintain high and stable performance across unseen sources.

\textbf{Time efficiency.} As illustrated in Figure \ref{efficiency}, both CoSur\textsubscript{LR} and CoSur significantly reduce the training time compared to LoRA-FT, while also greatly decreasing inference time relative to Base and ICL.

\begin{table}[t]
\centering
\setlength{\tabcolsep}{0.2mm}
\begin{tabular}{c|cccccc}
\toprule
Other & \multicolumn{2}{c}{CoSur$_{PCA}$} & \multicolumn{2}{c}{CoSur$_{CS}$} & \multicolumn{2}{c}{CoSur} \\
Source & ACC              & F1              & ACC             & F1              & ACC         & F1          \\ \midrule
Human                         & \underline{ 60.50}      & \underline{ 53.41}     & 14.75           & 13.11           & \textbf{100.00} & \textbf{100.00} \\
chatgpt                       & \underline{ 73.50}      & \underline{ 73.49}     & 5.25            & 5.24            & \textbf{99.75}  & \textbf{99.75}  \\
Llama                         & \underline{ 50.00}      & \underline{ 33.33}     & 11.75           & 11.63           & \textbf{99.25}  & \textbf{99.25}  \\
Deepseek                      & \underline{ 51.75}      & \underline{ 37.88}     & 8.25            & 8.12            & \textbf{97.00}  & \textbf{97.00}  \\ \midrule \rowcolor{blue!10}
Average                       & \underline{ 58.94}      & \underline{ 49.53}     & 10.00           & 9.53            & \textbf{99.00}  & \textbf{99.00}  \\ 
\bottomrule
\end{tabular}
\caption{Ablation study using Qwen.}
\label{variants}
\end{table}

\subsection{Ablation Study}
\label{ab}
The key to CoSur's performance lies in the construction of the subspace and the authorship discrimination based on projection energy. To evaluate the effectiveness of CoSur, we design two variants:
\textbf{(1) CS-based authorship identification (CoSur$_{CS}$):} For a given test sample, its authorship is determined by computing the cosine similarity between the sample and each class center.
\textbf{(2) PCA-based subspace construction (CoSur$_{PCA}$):} Subspaces are constructed using principal component analysis (PCA).
Compared to CoSur with a recognition accuracy of 99.00\%, CoSur$_{PCA}$ achieves significantly lower performance, with an average accuracy of 58.94\%. 
This indicates that the mean vectors constructed from texts generated by different models play an important role in distinguishing authorship. 
However, due to the anisotropy of LLM internal representations, the cosine similarity between mean vectors from different text sources remains high. 
As a result, CoSur$_{CS}$ achieves the lowest performance.

\begin{figure}[t]
\centering
\includegraphics[width=0.48\textwidth]{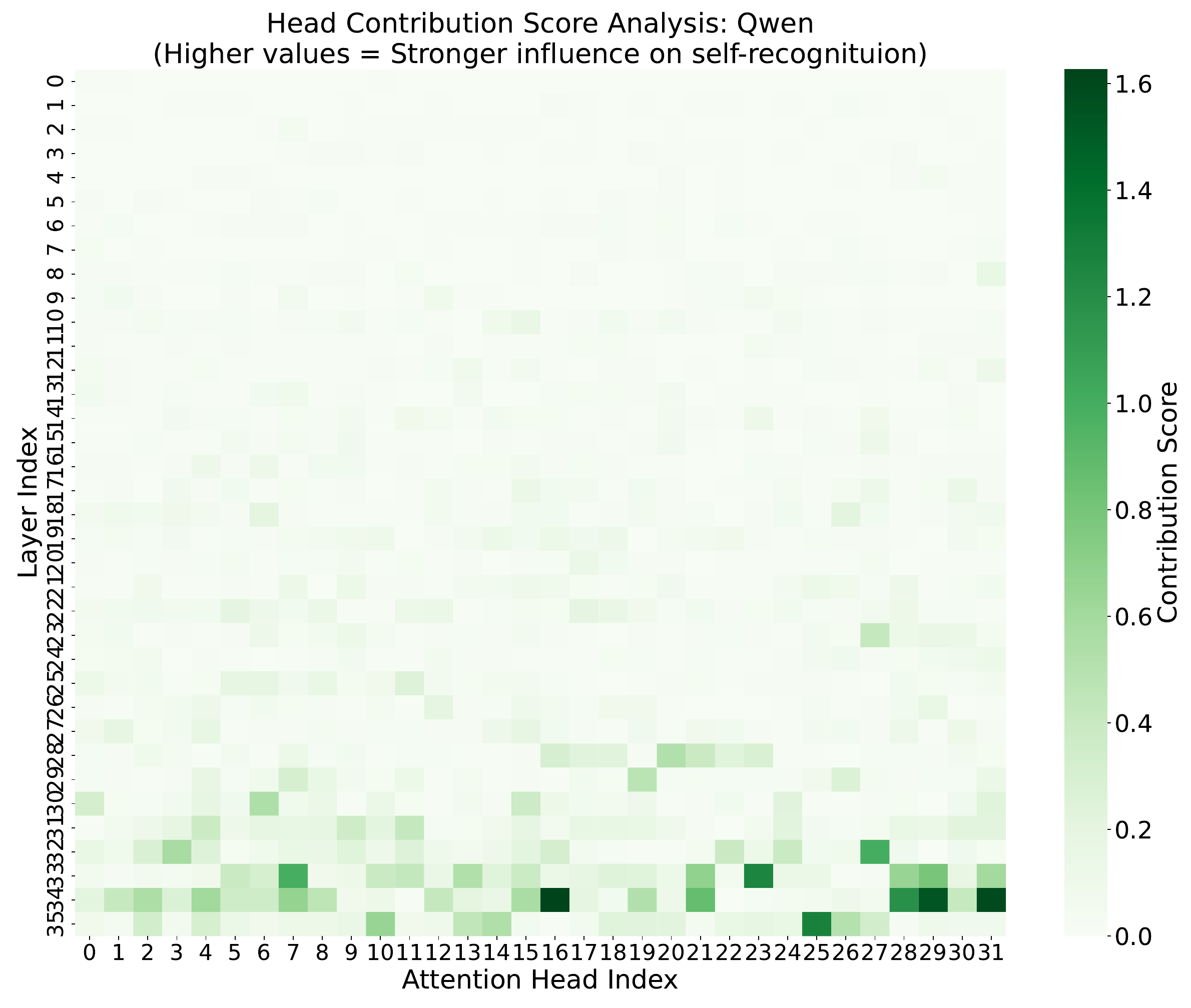} 
\caption{Heatmap of Qwen Attention Heads’ Self-Recognition Contribution Scores}
\label{contribution}
\end{figure}

\begin{figure}[t]
\centering
\includegraphics[width=0.48\textwidth]{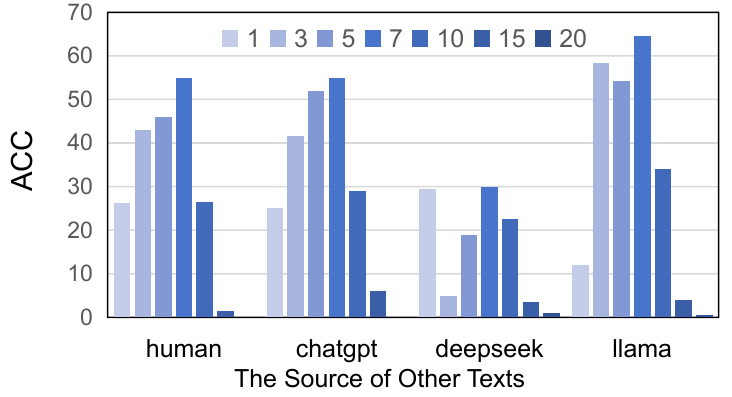} 
\caption{Effect of Amplifying Important Attention Heads with Different Scaling Factors on LLM Self-Recognition Accuracy under the IPP Scenario}
\label{AHI}
\end{figure}
To further examine whether the extracted self-recognition subspace captures information most relevant to self-recognition, we conduct attention-head intervention experiments.
For the $l_{th}$ layer and the $i_{th}$ attention head, we first compute its contribution to the LLM’s self-recognition ability $Score^{(i)}_l$, by projecting its output onto the self-recognition subspace:

\begin{equation}
Score^{(i)}_l
=
\frac{1}{B} \sum_{b=1}^{B} 
\Big\| \mathbf{x}_{b,i} \, (\mathbf{\mathcal{V}_s} \textbf{W}_O^{(i)})^\top \Big\|_2 \\
\end{equation}

\noindent
where $\textbf{W}_O^{(i)} \in \mathbb{R}^{d \times m}$ is the output projection matrix of this head. 
$\mathbf{x}_{b, i} \in \mathbb{R}^{m}$ denotes the post-attention activation of the \(i_{th}\) head in layer \(l\) for the \(b_{th}\) sample, computed as the weighted sum of the value vectors before applying the final linear projection $\textbf{W}_O^{(i)}$.
As shown in Figure \ref{contribution}, we observe the existence of attention heads within the LLM that are highly correlated with self-recognition.
We select the top 15 attention heads with the highest scores and amplify their outputs by different factors. 
As demonstrated in Figure \ref{AHI}, increasing the amplification factor can further enhance the LLM’s self-recognition accuracy under the IPP scenario.
However, excessive amplification can disrupt the model’s output behavior, leading to degraded classification performance, while insufficient amplification provides only limited improvement.
These results confirm that these heads encode information relevant to self-recognition and provide functional validation that the extracted subspace effectively captures such information.

We also investigate the performance of CoSur on a related task, LLM-generated text detection, as detailed in Appendix \ref{app:LLMGT}.

\begin{figure}[t]
\centering
\includegraphics[width=0.47\textwidth]{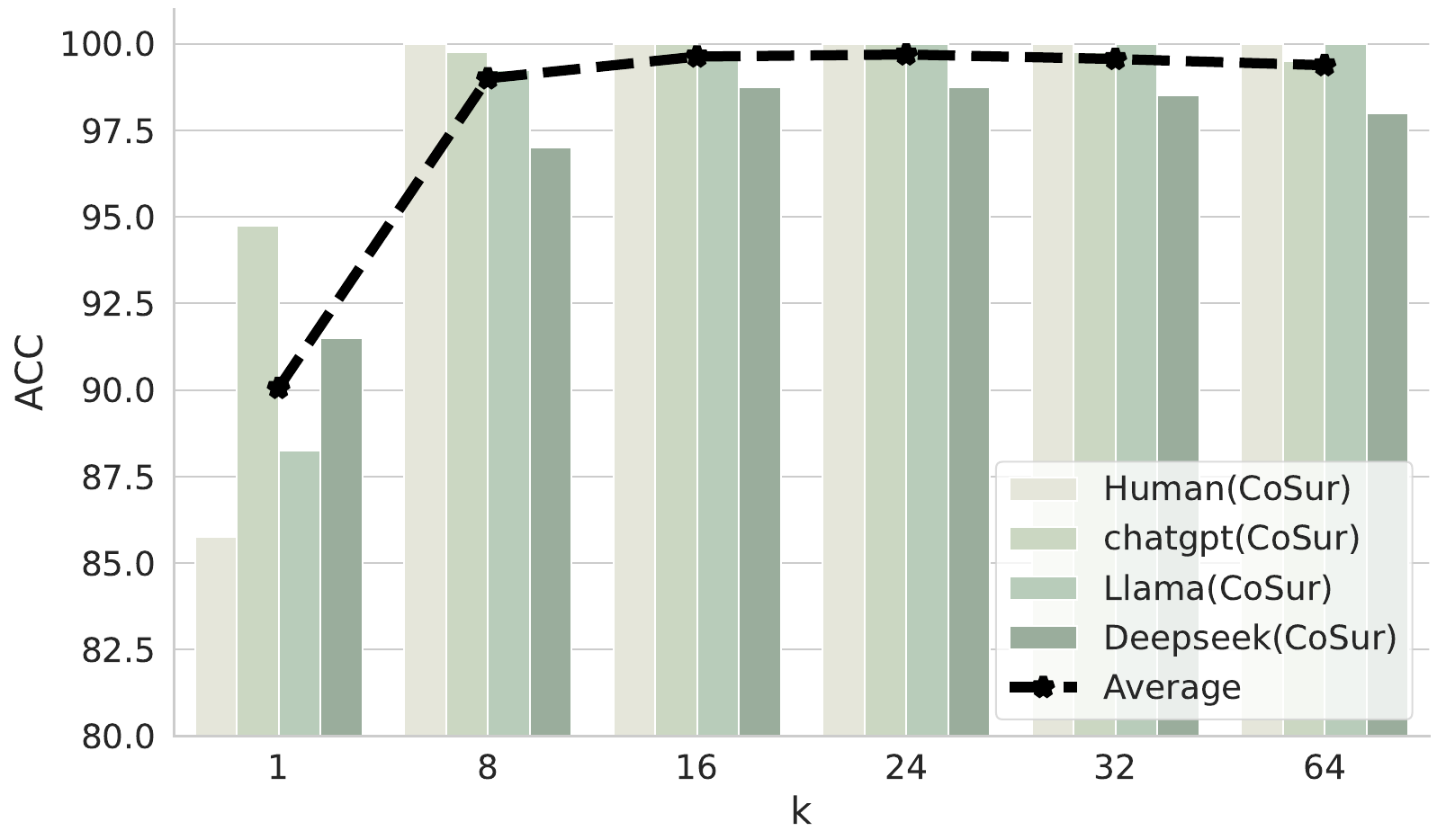} 
\caption{Accuracy variation with different Subspacedimensions using Qwen.}
\label{k_ab}
\end{figure}

\begin{figure}[t]
\centering
\includegraphics[width=0.45\textwidth]{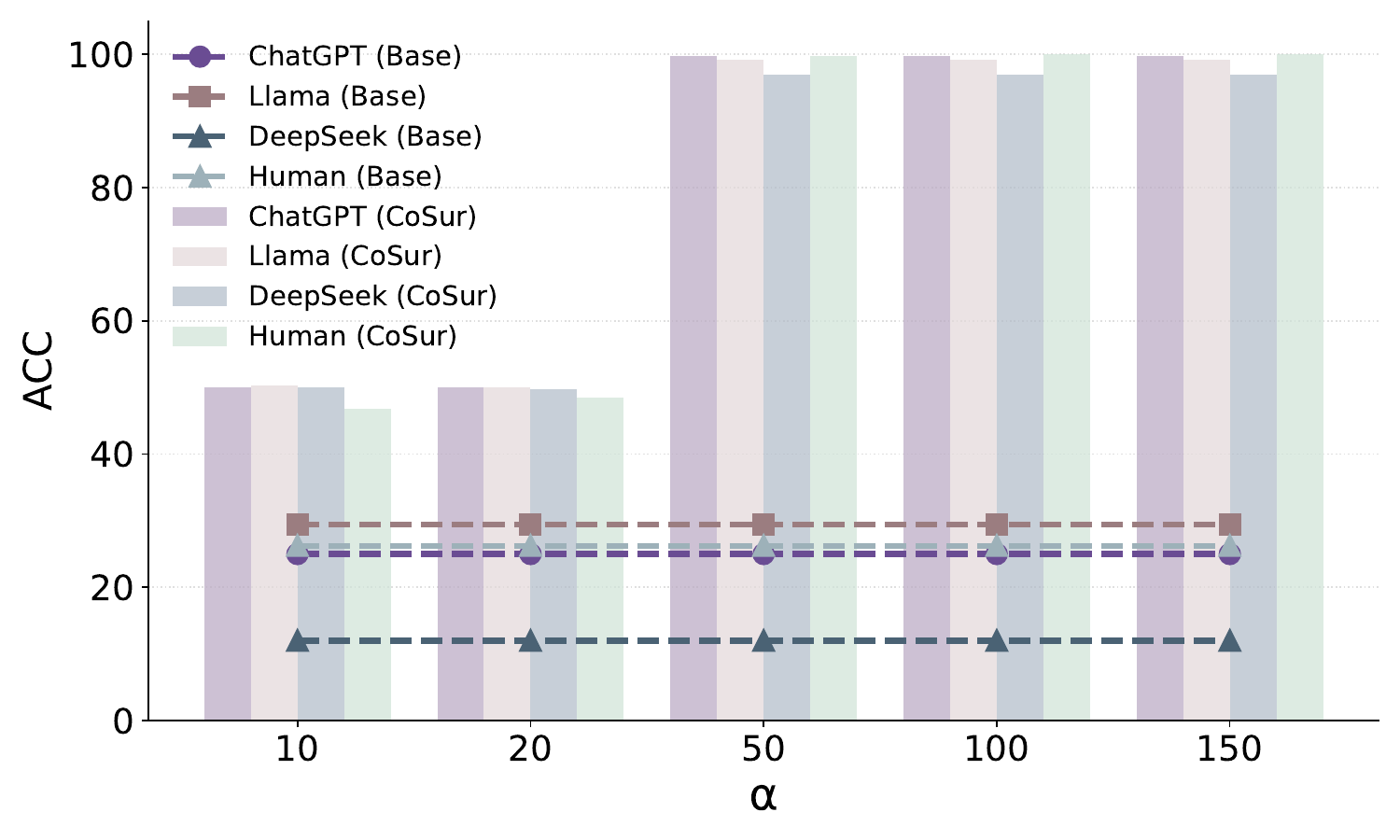} 
\caption{Accuracy variation with different editing Strength $\alpha$ using Qwen.}
\label{alpha_ab}
\end{figure}

\subsection{Hyperparameter Analysis}
We conduct hyperparameter analysis to study the sensitivity of our method to different settings.

\textbf{Impact of subspace dimensions $k$.} 
We explore the impact of different $k$ on the self-recognition ability of LLMs in the IPP scenario using Qwen.
As shown in Figure \ref{k_ab}, a small value of $k$ fails to capture all the distinguishing features. 
Increasing $k$ expands the subspace dimensionality, enabling the capture of more informative features and enhancing self-recognition performance.
When $k=8$, the average recognition accuracy reaches 99\%. 
Hence, we adopt $k=8$ in our experiments.

\textbf{The impace of editing strength $\alpha$.} 
Figure \ref{alpha_ab} illustrates that a small $\alpha$ fails to meaningfully affect the model's behavior, yielding only minor improvements in self-recognition performance. 
Once the editing strength surpasses 50\%, the average recognition accuracy across the four text combinations in IPP scenarios saturates, indicating that this threshold sufficiently guides the LLM to produce target-aligned responses.


\section{Conclusion}
In conclusion, we analyze the reasons behind the failure of this capability in the IPP scenario, which is regarded as the implicit self-recognition (ISR) of LLMs.
Based on this, we propose a novel method named cognitive surgery (CoSur) to mitigate the ISR in LLMs, enhancing LLMs’ performance in the IPP scenario. 
Experimental results demonstrate that CoSur significantly enhances LLMs' performance in the IPP scenario.

\section{Limitation}
The self-recognition subspace extracted in this study is intriguing, as it suggests that LLM-generated texts may exhibit model-specific representational patterns. 
However, the paper does not provide a detailed analysis of the properties of this subspace.
In future work, we will explore whether it reflects the distinctive stylistic features of specific LLMs and whether it can help attribute generated texts to their source models.



\bibliography{custom}

\appendix


\section{Detailed Prompt}
\label{app:prompt}

\begin{figure}
\centering
\includegraphics[width=0.45\textwidth]{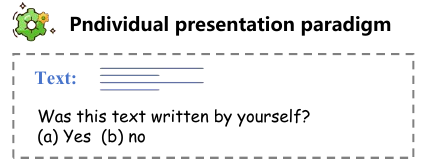} 
\caption{The detailed prompt used in the IPP scenario.}
\label{prompt}
\end{figure}

The detailed prompt used in the IPP scenario is provided in Figure \ref{prompt}.

\section{Analysis of the feature distributions}
\label{app:distribution}
We employ three metrics to quantify the similarity between different representation distributions: Mean Cosine Similarity (CS), Maximum Mean Discrepancy (MMD), and Linear Centered Kernel Alignment (CKA). Given two sets of representations $\{x_i\}_{i=1}^n \sim P$ and $\{y_j\}_{j=1}^m\sim Q$, their definitions are as follows:

\textbf{(1) Cosine Similarity.}
Given the mean embeddings  $\mu_x$ and $\mu_y$ of the two sets, we compute the cosine similarity as:
\begin{equation}
    CS(\mu_x, \mu_y) = \frac{\mu_x \cdot \mu_y}{\|\mu_x\| \, \|\mu_y\|}
\end{equation}
where $\mu_x = \frac{1}{n} \sum_{i=1}^{n} x_i$, $\mu_y = \frac{1}{m} \sum_{j=1}^{m} y_j$.
A higher value indicates greater alignment in the direction of the average feature vectors, suggesting stronger semantic similarity between the two text categories.

\textbf{(2) MMD.}
It measures the distributional distance between two sets of samples, and is defined as:

\begin{equation}
\begin{aligned}
    \mathrm{MMD}^2(P, Q) = & \mathbb{E}_{x,x'}[k(x, x')] + \mathbb{E}_{y,y'}[k(y, y')]\\ 
            & - 2 \mathbb{E}_{x,y}[k(x, y)]
\end{aligned}
\end{equation}
where $k( \cdot , \cdot )$ is a kernel function. 
In this work, we use the radial basis function (RBF) kernel:
\begin{equation}
k(x, y) = \exp\left(-\gamma |x - y|^2\right)
\end{equation}
Lower MMD values indicate that the two distributions are more similar in the feature space induced by the kernel, whereas higher values imply greater discrepancy.

\textbf{(3) CKA.}
Linear CKA measures structural similarity between representation matrices $X, Y \in \mathbb{R}^{n \times d}$, which is defined as:
\begin{equation}
    \mathrm{CKA}(X, Y) = 
\frac{\langle X X^\top, Y Y^\top \rangle_F}
{\|X X^\top\|_F \cdot \|Y Y^\top\|_F}
\end{equation}
where $\langle A, B \rangle_F = \sum_{i,j} A_{ij} B_{ij}$ denotes the Frobenius inner product and $\|\cdot\|_F$ denotes the Frobenius norm.
Higher CKA values indicate stronger similarity in the relational structure of the two representations.

We select 600 samples from each text category and analyze the differences in feature distributions extracted from Llama and Deepseek.
As shown in Table \ref{feature_distance}, similar to Qwen, both the features of different categories in Llama and Deepseek exhibit high CS and MMD, but low CKA. 
This indicates that while these features demonstrate high semantic similarity, they possess structural heterogeneity. 
These observations demonstrate that it is possible to identify and extract subspaces where the representations exhibit more pronounced differences, thereby enhancing inter-category discriminability.
However, compared to Qwen and Deepseek, Llama exhibits higher CKA values, indicating weaker feature discriminability. We attribute this phenomenon to feature entanglement caused by its Grouped-Query Attention (GQA) architecture.

\begin{table}[t]
\centering
\setlength{\tabcolsep}{0.8mm}
\begin{tabular}{ccccc}
\toprule
     LLM & Other Source & CS & MMD & CKA  \\
\midrule
    \multirow{4}{*}{Llama} &Human & 0.9283 & 0.0340 & 0.1651 \\
    &ChatGPT & 0.9614 & 0.0036 & 0.0687 \\ 
    &Qwen & 0.8624 & 0.0048 & 0.1497 \\
    &Deepseek & 0.8857 & 0.0041 & 0.1184 \\
    \midrule
    \multirow{4}{*}{Deepseek} &human & 0.9303 & 0.0033 & 0.0740 \\
    &ChatGPT & 0.9752 & 0.0034 & 0.0485 \\ 
    &Qwen & 0.9921 & 0.0034 & 0.0379 \\
    &Llama & 0.9839 & 0.0034 & 0.0227 \\
\bottomrule
\end{tabular}
\caption{The distance between feature representations of texts generated by different models in Llama and Deepseek.
The Other-source column indicates the model used to generate other texts.
}
\label{feature_distance}
\end{table}

\section{Subspace Spatial Distance Metric}
\label{app: subspace}
To evaluate the representational capacity and separability of the subspaces we constructed, we compute the Normalized Grassmann Distance (NGD) and Normalized Frobenius Distance(NFD) between them.
Given two subspaces with orthonormal basis matrices $U, V \in \mathbb{R}^{d \times k}$, the NGD and NFD is defined as follows:

\textbf{(1) NGD.}
It quantifies the geometric distance between two subspaces based on their principal angles, which is defined as:
\begin{equation}
    \mathrm{NGD}(U, V) = \frac{1}{\sqrt{k} \cdot \frac{\pi}{2}} \sqrt{\sum_{i=1}^k \theta_i^2},
\end{equation}
where $\theta_i = \arccos(\sigma_i)$ and $\sigma_i$ are the singular values of the matrix $ U^\top V $.

\textbf{(2) NFD.}  
It measures the difference between the projection matrices of the two subspaces, normalized by its maximum value:
\begin{equation}
    \mathrm{NFD}(U, V) = \frac{\| U U^\top - V V^\top \|_F}{\sqrt{2k}},
\end{equation}
where $\|\cdot\|_F$ denotes the Frobenius norm. Higher values of NGD and NFD indicate greater dissimilarity between the two subspaces.

\begin{table}
\centering
\setlength{\tabcolsep}{1mm}
\begin{tabular}{cccc}
\toprule
     Target LLM & Other Source & NGD & NFD   \\
\midrule
    \multirow{4}{*}{Llama}&Human & 0.6413 & 0.7717\\
    &ChatGPT & 0.5815 & 0.7168 \\
    &Deepseek & 0.5853 & 0.7248 \\
    &Qwen  & 0.5961 & 0.7366 \\  \midrule
    \multirow{4}{*}{Deepseek}&Human & 0.6852 & 0.8232\\
    &ChatGPT & 0.6508 & 0.7915 \\
    &Qwen & 0.5902 & 0.7316 \\
    &Llama  & 0.6313 & 0.7723 \\  
\bottomrule
\end{tabular}
\caption{Measurement of distances between different subspaces using Qwen.
The self-recognition subspace is constructed from Qwen-texts, and the Other-source column indicates the model used to generate texts for constructing the other-recognition subspace.
}
\label{subspace_distance_multiLLM}
\end{table}

As shown in Table \ref{subspace_distance_multiLLM}, the NGD and NFD between these subspaces are large, indicating significant divergence between their corresponding subspaces.

\section{Algorithm of CoSur}
\label{app:algorithm}
We first construct a subspace for each text category, including self-generated texts and other-generated texts, using SVD. 
The construction process is detailed in Algorithm \ref{alg:subspace}.

Given the constructed subspace subspaces, we perform authorship discrimination based on projection energy and cognitive editing to modify the last-token hidden representation $h$ in the final layer of the target LLM into an edited representation $\tilde{h}$, thereby guiding the LLM toward producing a correct response.
The full algorithm is presented in Algorithm \ref{alg:editing}.

\begin{algorithm}[t]
\caption{SVD-Based subspace Construction}
\label{alg:subspace}
\begin{algorithmic}[1]
\REQUIRE Feature matrix $\mathbf{H} \in \mathbb{R}^{N \times d}$, target rank $k$
\ENSURE subspace basis $\mathcal{V} \in \mathbb{R}^{d \times k}$
\STATE Perform truncated SVD: $\mathbf{H} \approx \mathbf{U} \mathbf{\Sigma} \mathbf{V}^\top$
\STATE Extract top-$k$ right singular vectors: $\mathcal{V} \gets \mathbf{V}^{(k)}$
\RETURN $\mathcal{V}$
\end{algorithmic}
\end{algorithm}

\begin{algorithm}[t]
\caption{CoSur}
\label{alg:editing}
\begin{algorithmic}[1]
\REQUIRE Last token hidden states $h \in \mathbb{R}^{d}$, edit strength $\alpha$, self-subspace basis $\mathcal{V}_s \in \mathbb{R}^{d \times k}$, Other-subspace basis $\mathcal{V}_o \in \mathbb{R}^{d \times k}$, weight vector of target token for self-generated text $\mathbf{w}_s$,  weight vector of target token for other-generated text $\mathbf{w}_o$
\ENSURE Perturbed hidden states $H'$
\STATE Get target direction:
$\tilde{\mathbf{w}}_o= \frac{\mathbf{w}_o}{\|\mathbf{w}_o\|}$, 
$\tilde{\mathbf{w}}_s = \frac{\mathbf{w}_s}{\|\mathbf{w}_s\|}$
    \STATE Compute projection energies onto basis $E_s$ and $E_o$
    \IF{$E_s > E_o$}
        \STATE $\tilde{h} \gets h + \alpha \cdot \tilde{\mathbf{w}}_s$
    \ELSE
        \STATE $\tilde{h} \gets h + \alpha \cdot \tilde{\mathbf{w}}_o$
    \ENDIF
\RETURN $\tilde{h}$
\end{algorithmic}
\end{algorithm}

\begin{figure*}
\centering
\includegraphics[width=1\textwidth]{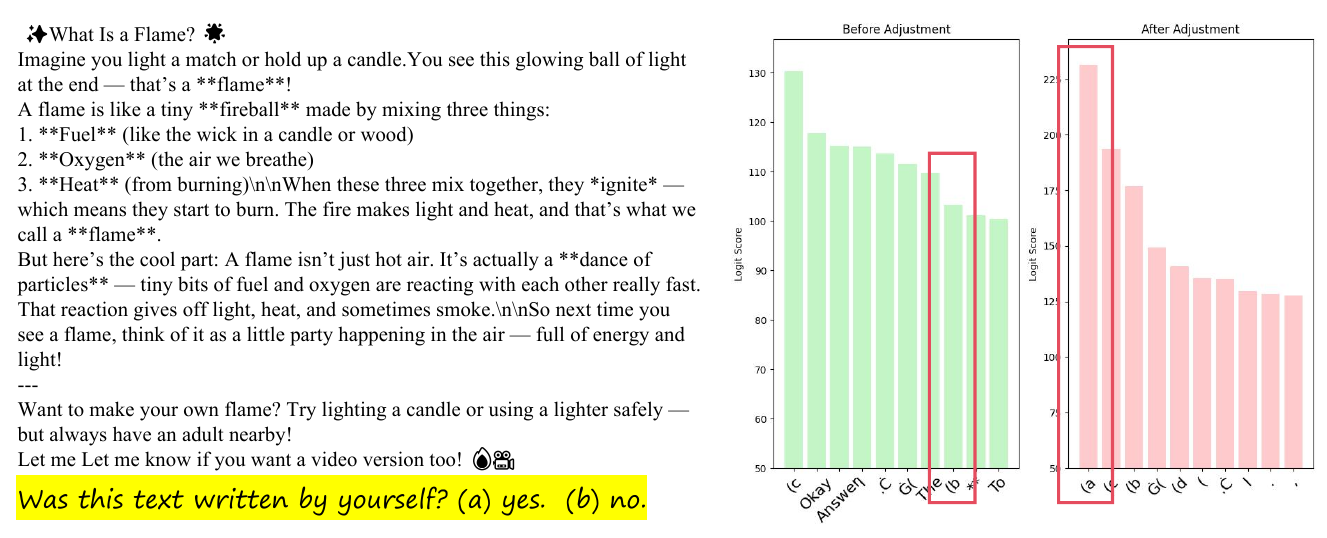} 
\caption{A Case Study on Applying CoSur to Influence Qwen's Outputs.}
\label{case_study}
\end{figure*}

\section{Additional Evaluation Using Fail Negative Rate (FNR) }
\label{app: additional_exp}
The Fail Negative Rate (FNR) measures the proportion of self-generated texts that are incorrectly classified as other-generated. Formally, it is defined as:

\begin{equation}
    \text{FNR} = \frac{1}{B} \sum_{b=1}^{B} \mathbf{1} \big[ \hat{y}(\mathbf{t}_i) = \text{other} \big],
\end{equation}
where $B$ is the number of self-generated texts in the evaluation batch, $\mathbf{1}[\cdot]$ denotes the indicator function, which equals 1 if the condition inside is true and 0 otherwise.
$\hat{y}(\mathbf{t}_b)$ is the predicted label of the $b_{th}$ self-generated text $\mathbf{t}_b$, taking values in $\{\text{self}, \text{other}\}$.
As shown in Table \ref{FNR}, we report the Fail Negative Rate (FNR) across different methods and text sources on Qwen under the IPP scenario.
Baseline and ICL exhibit consistently high FNR, indicating frequent failures in recognizing self-generated texts.
In contrast, CoSur and CoSur$_{LR}$ reduce FNR to near-zero levels, demonstrating substantially improved reliability in self-recognition.

\begin{table}
\centering
\setlength{\tabcolsep}{0.5mm}
\begin{tabular}{cccccc}
\toprule
         & Base  & ICL   & LoRa-FT & CoSur$_{LR}$ & CoSur \\ \midrule
Human    & 80.50 & 23.00 & 0.00    & 5.50    & 0.00  \\
ChatGPT  & 73.50 & 35.50 & 0.00    & 1.50    & 0.50  \\
Llama    & 77.00 & 42.50 & 0.00    & 0.00    & 1.00  \\
Deepseek & 80.00 & 35.00 & 0.50    & 0.50    & 1.00 \\
\bottomrule
\end{tabular}
\caption{Fail Negative Rate (FNR) across different methods and text sources using Qwen under the IPP scenario. }
\label{FNR}
\end{table}

\section{Case Study}
\label{app:case}
Figure \ref{case_study} illustrates the changes in the Qwen model's top-10 output tokens and their logits before and after applying CoSur. 
The results demonstrate that CoSur substantially improves the model’s accuracy in text authorship attribution under the IPP scenario.

\begin{table}[t]
\centering
\setlength{\tabcolsep}{0.5mm}
\begin{tabular}{cccc|c}
\toprule
      & ChatGPT & Qwen & Deepseek & Average  \\
\midrule
   ChatGPT-D &\textbf{0.9950} & 0.7325 & 0.6475 & 0.6333 \\
   Fast-D & 0.8125 & 0.7175 & 0.3700 & 0.7917 \\ 
   Binoculars & 0.9625 & 0.8550 & 0.4775 & 0.6175 \\ 
   CoSur$_{Qwen}$& 0.8471 & \textbf{0.8571} & \textbf{0.8571} & \textbf{0.8538} \\
\bottomrule
\end{tabular}
\caption{Accuracy on the LLM-GT detection task.
}
\label{aigt}
\end{table}


\section{Performance on LLM-generated text detection}
\label{app:LLMGT}
As powerful tools for streamlining content creation, LLMs are widely used across various domains, including journalism, academia, and social media. 
However, the threats posed by LLM-generated text (LLM-GT), such as academic dishonesty, fake news, and false comments have raised significant concerns. 
To prevent the LLM abuse with malicious purpose, numerous LLM-GT detection methods have been proposed.
The goal of LLM-GT detection task is to distinguish between AI-generated texts and human-written texts.

Recent study \cite{bhattacharjee2024fighting} have shown that directly using LLMs as detectors is unreliable. 
A simple approach to improve LLM performance on LLM-GT detection tasks is to fine-tune the model. 
However, fine-tuning LLMs requires significant resource consumption, making it impractical.
Therefore, applying our proposed CoSur, which is a training-free method, to this task is both urgent and practically significant. 
We compare CoSur with existing state-of-the-art LLM-GT detection methods on the dataset used in our study. All models that require training are trained on the ChatGPT-human dataset, and the subspace for CoSur is constructed based on ChatGPT and human text.

The baseline methods we used are introduced as follows:
\textbf{(1) ChatGPT-detector (ChatGPT-D)} \cite{guo2023closeChatGPThumanexperts} is a Roberta-based model fine-tuned on the text generated by GPT-3.5 \cite{ouyang2022traininglanguagemodelsfollow}.
\textbf{(2) Fast-DetectGPT (Fast-D)} \cite{bao2024fastdetectgpt} is an optimized zero-shot detector, which utilize conditional probability curvature to elucidate discrepancies in word choices between LLMs and humans within a given context. In this study, both the sampling model and the scoring model used in the method are GPT-2 \cite{radford2019language}.
\textbf{(3) Binoculars} \cite{hans2024spotting} detects AI text by measuring the ratio between an observer model's perplexity and its cross-perplexity on a performer model's outputs.

As shown in Table \ref{aigt}, CoSur outperforms existing SOTA methods, achieving an average accuracy of  demonstrating its significant ability to enhance LLM performance in LLM-GT detection tasks.

\section{LLM Usage Statement}
In preparing this manuscript, we use GPT-5 \cite{GPT-5} for language polishing and stylistic refinement.
All scientiffc content, experimental design, analysis, and conclusions were independently
developed by the authors.


\end{document}